\documentclass[11pt,a4paper]{article}
\usepackage{times,latexsym}
\usepackage{url}
\usepackage[T1]{fontenc}

\usepackage[acceptedWithA]{tacl2018v2}

\newif\iftaclinstructions
\taclinstructionsfalse 
\iftaclinstructions
\renewcommand{\confidential}{}
\renewcommand{\anonsubtext}{(No author info supplied here, for consistency with
TACL-submission anonymization requirements)}
\newcommand{\instr}
\fi

\iftaclpubformat 

\else

\fi

\title{Learning Multilingual Word Embeddings in\\ Latent Metric Space: A Geometric Approach}

\author{
 Pratik Jawanpuria$^{1}$, Arjun Balgovind$^{2}$\Thanks{This work was carried out during the author's internship at Microsoft, India.} , Anoop Kunchukuttan$^{1}$, Bamdev Mishra$^{1}$\\
 {}$^{1}$Microsoft, India \quad\quad {}$^{2}$IIT Madras, India \\
  {}$^{1}$\texttt{\{pratik.jawanpuria,ankunchu,bamdevm\}@microsoft.com}\\
  {}$^{2}$\texttt{barjun@cse.iitm.ac.in}\\
}

\date{}

\usepackage{enumitem}
\setlist{nolistsep}
\usepackage{nicefrac}
\usepackage{booktabs}
\usepackage{mathtools}
\usepackage{color,multirow,wrapfig}
\usepackage[normalem]{ulem}
\usepackage{breakurl} 
\usepackage{graphicx} 
\usepackage{epsfig} 
\usepackage{subfigure} 

\usepackage{algorithm}
\usepackage{algorithmic}

\usepackage{amssymb}
\usepackage{amsmath}
\usepackage{mathtools}

\newcommand{\ie}{\textit{, i.e., }}

\newcommand{\thead}[1]{\multicolumn{1}{c}{\textbf{#1}}}
\newcommand{\lhead}[1]{\multicolumn{1}{l}{\textbf{#1}}} 

\newcommand{\algname}{\mbox{GeoMM}}

\newcommand{\bulletpoint}{\noindent$\bullet$\;\;}

\def\R{\mathbb{R}}

\def\O{\mathbb{O}}

\def\bzero{{\mathbf 0}}

\def\bB{\mathbf B}
\def\bI{{\mathbf I}}

\def\bU{{\mathbf U}}
\def\bV{{\mathbf V}}
\def\bW{{\mathbf W}}
\def\bX{{\mathbf X}}
\def\bY{{\mathbf Y}}

\def\minop{\mathop{\rm min}\limits}

\begin{document} 
\maketitle





\begin{abstract} 


We propose a novel geometric approach for learning bilingual mappings given monolingual embeddings and a bilingual dictionary. Our approach  {decouples} the source-to-target language transformation into (a) language-specific rotations on the original embeddings to align them in a common, latent space, and (b) a  language-independent similarity metric in this common space to better model the similarity between the embeddings. Overall, we pose the bilingual mapping problem {as a classification problem} on smooth Riemannian manifolds. Empirically, our approach outperforms previous approaches on the bilingual lexicon induction and cross-lingual word similarity tasks. 

We next generalize our framework to represent multiple languages in a common latent space. Language-specific rotations for all the languages and a common similarity metric in the latent space are learned \textit{jointly} from bilingual dictionaries for multiple language pairs. We illustrate the effectiveness of joint learning for multiple languages in an indirect word translation setting. 
\end{abstract}

\section{Introduction}\label{sec:intro}

Bilingual word embeddings are a useful tool in NLP that has attracted a lot of interest lately, due to a fundamental property: similar concepts/words across different languages are  mapped close to each other in a common embedding space. Hence, they are useful for joint/transfer learning and sharing annotated data across languages in different NLP applications like machine translation \citep{gu2018}, building bilingual dictionaries \citep{mikolov13a}, mining parallel corpora \citep{conneau18a}, text classification \citep{klementiev12a}, sentiment analysis \citep{zhou15}, and dependency parsing \citep{ammar16}. 

\citet{mikolov13a} empirically show that a linear transformation of embeddings from one language to another preserves the geometric arrangement of word embeddings. 
In a supervised setting, the transformation matrix, $\bW$, is learned given a small bilingual dictionary and their corresponding monolingual embeddings. 
Subsequently, many refinements to the bilingual mapping framework  have been proposed. \citep{xing15a,smith17a,conneau18a,artetxe16a,artetxe17a,artetxe18a,artetxe18b}. 


In this work, we propose a novel geometric approach for learning bilingual embeddings. We rotate the source and target language embeddings from their original vector spaces to a common latent space via language-specific orthogonal transformations. Furthermore, we define a similarity metric, the Mahalanobis metric, in this common space to refine the notion of similarity between a pair of embeddings. We achieve the above by learning the transformation matrix as follows: $\bW=\bU_{t}\bB\bU^\top_{s}$, where $\bU_t$ and $\bU_s$ are the orthogonal transformations for target and source language embeddings, respectively,  and $\bB$ is a positive definite matrix representing the Mahalanobis metric. 

The proposed formulation has the following benefits:

\noindent\bulletpoint The learned similarity metric allows for a more effective similarity comparison of embeddings based on evidence from the data.

\noindent\bulletpoint A common latent space decouples the source and target language transformations, and naturally enables representation of  word embeddings from both languages in a single vector space. 

\noindent\bulletpoint We also  show that the proposed method can be  easily generalized to jointly learn multilingual embeddings, given bilingual dictionaries of multiple language pairs. We map multiple languages into a single vector space by learning the characteristics common across languages (the similarity metric) as well as language specific attributes (the orthogonal transformations).

The optimization problem resulting from our formulation  involves orthogonal constraints on language-specific transformations ($\bU_i$ for language $i$) as well as the symmetric positive-definite constraint on the metric $\bB$. Instead of solving the optimization problem in the Euclidean space with constraints, we view it as an optimization problem in smooth Riemannian manifolds, which are well-studied topological spaces \citep{lee03a}. The Riemannian optimization framework embeds the given constraints into the search space, and conceptually views the problem as an {unconstrained} optimization problem over the manifolds. 

We evaluate our approach on different bilingual as well as multilingual tasks across multiple languages and datasets. The following is a summary of our findings: 

\bulletpoint  Our approach outperforms state-of-the-art supervised and unsupervised bilingual mapping methods on the bilingual lexicon induction as well as the cross-lingual word similarity tasks.

\bulletpoint An ablation analysis reveals that the following contribute to our model's improved performance: (a) aligning the embedding spaces of different languages, (b) learning a similarity metric which induces a latent space, (c) {performing inference in the induced latent space}, and (d) formulating the tasks as a classification problem. 

\bulletpoint We evaluate our multilingual model on an indirect word translation task: translation between a language pair that does not have a bilingual dictionary, but the source and target languages each possess a  bilingual dictionary with a third, common pivot language. Our multilingual model outperforms a strong unsupervised baseline as well as methods based on adapting bilingual methods for this indirect translation task.

\bulletpoint Lastly, we propose a semi-supervised extension of our approach which further improves performance over the supervised approaches.

The rest of the paper is organized as follows. Section~\ref{sec:related} discusses related work. The proposed framework, including problem formulations for bilingual and multilingual mappings, is presented in Section~\ref{sec:polar_factorization}. The proposed Riemannian optimization algorithm is described in Section~\ref{sec:optimization}. In Section~\ref{sec:exp_setup}, we discuss our experimental setup. Section~\ref{sec:dir_trans} presents the results of experiments on direct translation with our algorithms and analyzes the results. Section~\ref{sec:indir_trans} presents experiments on indirect translation using our generalized multilingual algorithm. We discuss a semi-supervised extension to our framework in Section~\ref{sec:semi_sup}. Section \ref{sec:conclusion} concludes the paper. 

\section{Related Work}
\label{sec:related}



\textbf{Bilingual embeddings}. \citet{mikolov13a} show that a linear transformation from embeddings of one  language to another can be learned from a bilingual dictionary and corresponding monolingual embeddings by performing linear least-squares regression. A popular modification to this formulation constrains the transformation matrix to be orthogonal \citep{xing15a,smith17a,artetxe18a}. This is known as the  \textit{orthogonal Procrustes  problem} \citep{schonemann1966a}. Orthogonality preserves monolingual distances and ensures the transformation is reversible.  
\citet{lazaridou15a} and \citet{joulin18b} optimize alternative loss functions in this framework.
\citet{artetxe18a} improves upon the Procrustes solution and propose a multi-step framework consisting of a series of linear transformations to the data. \citet{faruqui14a} use {Canonical Correlation Analysis} (CCA) to learn linear projections from the source and target languages to a common space such that correlations between the embeddings projected to this space are maximized. Procrustes solution based approaches have been shown to perform better than CCA-based approaches \citep{artetxe16a,artetxe18a}. 

We view the problem of mapping the source and target languages word embeddings as (a) aligning the two language spaces, and (b) learning a similarity metric in this (learned) common space. We accomplish this by learning suitable language-specific orthogonal transformations (for alignment) and a symmetric positive-definite matrix (as Mahalanobis metric). 
The similarity metric is useful in addressing the limitations of mapping to a common latent space under orthogonality constraints, an issue discussed by \citet{doval18a}. While \citet{doval18a} learn a second correction transformation by assuming the average of the projected source and target embeddings as the true latent representation, we make no such assumption and learn the similarity metric from the data. 
\citet{kementchedjhieva18a}, recently, employed the generalized Procrustes analysis (GPA) method \citep{gower75} for the bilingual mapping problem. GPA maps both the source and target language embeddings to a latent space, which is constructed by averaging over the two language spaces. 



Unsupervised methods have shown promising results, matching supervised methods in many studies. \citet{artetxe17a} proposed a bootstrapping method for bilingual lexicon induction problem using a small seed bilingual dictionary. Subsequently, \citet{artetxe18b} and \citet{hoshen18} have proposed initialization methods that eliminate the need for a seed dictionary. \citet{zhang17b} and \citet{grave18}  proposed aligning the the source and target language word embeddings by optimizing the the Wasserstein distance. Unsupervised methods based on adversarial training objectives have also been proposed \citep{barone16,zhang17a,conneau18a,chen18a}. A recent work by \citet{sogaard18} discusses cases in which unsupervised bilingual lexicon induction does not lead to good performance. 

\noindent\textbf{Multilingual embeddings}.  
\citet{ammar16} and \citet{smith17b} adapt bilingual approaches for  representing embeddings of multiple languages in a common vector space by designating one of the languages as a \textit{pivot} language. In this simple approach, bilingual mappings are learned {\it independently} from all other languages to the pivot language.  
GPA based method \citep{kementchedjhieva18a} may also be used to jointly transform multiple languages to a common latent space. 
However, this  requires an $n$-way dictionary to represent $n$ languages. In contrast, the proposed approach requires only pairwise bilingual dictionaries such that every  language under consideration is represented in at least one bilingual dictionary.


The above-mentioned approaches are referred to as \textit{offline} since the monolingual and bilingual embeddings are learned separately. In contrast \textit{online} approaches directly learn a bilingual/multilingual embedding from parallel corpora \citep{herman14a,huang15a,duong17a}, optionally augmented with monolingual corpora \citep{klementiev12a,chandar14a,gouws15a}. In this work, we focus on offline approaches.

\section{Learning Latent Space Representation}
\label{sec:polar_factorization}

In this section, we first describe the proposed geometric framework to learn bilingual embeddings. We then present its generalization  to the multilingual setting. 

\subsection{Geometry-aware Factorization}\label{subsec:geometry}

We propose to transform the word embeddings from the source and target languages to a common space in which the similarity of words embeddings may be better learned. To this end, we \textit{align} the source and target languages embedding spaces by learning language-specific rotations: $\bU_s\in\O^d$ and $\bU_t\in\O^d$ for the source and target languages embeddings, respectively. Here $\O^d$ represents the space of $d$-dimensional orthogonal matrices. An embedding $x$ in the source language is thus transformed to $\psi_s(x)=\bU_s^\top x$. Similarly, for an embedding $z$ in the target language: $\psi_t(z)=\bU_t^\top z$. 
These orthogonal transformations map (align) both the source and target language embeddings to a common space in which we learn a data-dependent similarity measure, as discussed below. 

We learn a Mahalanobis metric $\bB$ to {\it refine} the notion of similarity\footnote{Mahalanobis metric generalizes the notion of cosine similarity. For given two unit normalized vectors $x_1,x_2\in\R^d$, their cosine similarity is given by $\mathrm{sim}_{\bI}(x_1,x_2)=x_1^\top\bI x_2=x_1^\top x_2$, where $\bI$  is the identity matrix. If this space is endowed with a metric $\bB\succ\bzero$, then $\mathrm{sim}_{\bB}(x_1,x_2)=x_1^\top\bB x_2$. } between
the two transformed embeddings $\psi_s(x)$ and $\psi_t(z)$. 
The Mahalanobis metric incorporates the feature correlation information from the given training data. This allows for a more effective similarity comparison of language embeddings (than the cosine similarity). 
In fact, Mahalanobis similarity measure reduces to cosine similarity when the features are uncorrelated and have unit variance, which may be a strong assumption in real-world applications. \citet{sogaard18} have argued that monolingual embedding spaces across languages are not necessarily isomorphic, hence learning a orthogonal transformation alone may not be sufficient. A similarity metric learned from the data may mitigate this limitation to some extent by learning a correction in the latent space.

Since $\bB$ is a Mahalanobis metric in $\R^d$ space, it is a $d\times d$ symmetric positive-definite matrix $\bB$\ie $\bB\succ\bzero$. 
The similarity between the embeddings $x$ and $z$ in the proposed setting is computed as  $h_{st}(x,z)=\psi_t(z)^\top \bB \psi_s(x)=z^\top (\bU_t \bB \bU_s^\top) x$. The source to the target language transformation is expressed as  $\bW_{ts}=\bU_t \bB \bU_s^\top$. 
For an embedding $x$ in the source language, its transformation to the target language space is given by $\bW_{ts}x$. 

The proposed factorization of the transformation $\bW=\bU\bB\bV^\top$, where $\bU,\bV\in\O^d$ and $\bB\succ\bzero$, is sometimes referred to as polar factorization of a matrix \citep{bonabel10,meyer11a}. 
Polar factorization is similar to the singular value decomposition (SVD) The key difference is that SVD enforces $\bB$ to be a {\it diagonal} matrix with non-negative entries, which accounts for only the axis rescaling instead of full feature correlation and is more difficult to optimize \citep{mishra14,harandi17}.

\subsection{Latent Space Interpretation} \label{subsec:latentspace}

Computing the Mahalanobis similarity measure is equivalent to computing the cosine similarity in a special latent (feature) space. This latent space is defined by the transformation $\phi:\R^d\rightarrow\R^d$, where the mapping is defined as $\phi(w)=\bB^{\frac{1}{2}} w$. Since $\bB$ is a symmetric positive-definite matrix, $\bB^{\frac{1}{2}}$ is well-defined and unique. 


Hence, our model may equivalently be viewed as learning a suitable latent space as follows. The source and target languages embeddings are linearly transformed as $x\mapsto\phi(\psi_s(x))$ and $z\mapsto\phi(\psi_t(z))$, respectively. The  functions $\phi(\psi_s(\cdot))$ and $\phi(\psi_t(\cdot))$ map the source and target language embeddings, respectively, to a common latent space. We learn the matrices $\bB$, $\bU_s$, and $\bU_t$ corresponding to the transformations $\phi(\cdot)$, $\psi_s(\cdot)$, and $\psi_t(\cdot)$, respectively. Since the matrix $\bB$ is embedded implicitly in this latent feature space, we employ the usual cosine similarity measure, computed as $\phi(\psi_t(z))^\top \phi(\psi_s(x))=z^\top \bU_t \bB \bU_s^\top x$. It should be noted that this is equal to $h_{st}(x,z)$. 
 


\subsection{A Classification Model}\label{subsec:bilingual}
We assume a small bilingual dictionary (of size $n$) is available as the training data. 
Let $\bX_s\in\R^{d\times n_s}$ and $\bX_t\in\R^{d\times n_t}$ denote the embeddings of the dictionary words from the source and target languages, respectively. Here, $n_s$ and $n_t$ are the number of unique words in the source and target languages present in the dictionary.



We propose to model the bilingual word embedding mapping problem as a binary classification problem. 
Consider word embeddings $x$ and $z$ from the source and target languages, respectively. If the words corresponding to $x$ and $z$ constitute a translation pair then the pair $\{x,z\}$ belongs to the positive class, else it belongs to the negative class. The prediction function for the pair $\{x,z\}$ is $h_{st}(x,z)$. 
We create a binary label matrix $\bY_{st}\in\{0,1\}^{n_s\times n_t}$ whose $(i,j)$-{th} entry corresponds to the correctness of mapping the $i$-{th} embedding in $\bX_s$ to the $j$-{th} embedding in $\bX_t$. 
Our overall optimization problem is as follows: 
\begin{flalign}\label{eqn:bilingualsquareloss}
\minop_{\bU_s\in\O^d,\bU_t\in\O^d,\bB\succ\bzero}&\  \|\bX_s^\top\bU_s\bB\bU_t^\top \bX_t-\bY_{st}\|_F^2\nonumber\\
 & + \lambda \|\bB\|_F^2. 
\end{flalign}
where $\|\cdot\|_F$ is the Frobenius norm and $\lambda>0$ is the regularization parameter. We employ the square loss function since it is smooth and relatively easier to optimize. 
It should be noted that our prediction function is invariant of the direction of mapping\ie $h_{st}(x,z)=h_{ts}(z,x)$. Hence, our model learns bidirectional mapping. The transformation matrix from the target to the source language is  given by $\bW_{st}=\bU_s \bB \bU_t^\top$\ie $\bW_{st}=\bW_{ts}^\top$.

%
%

The computation complexity of computing the loss term in (\ref{eqn:bilingualsquareloss}) is linear in $n$, the size of the given bilingual dictionary. This is because the loss term in (\ref{eqn:bilingualsquareloss}) can be re-written as follows: 
\begin{flalign}
&\|\bX_s^\top\bU_s\bB\bU_t^\top \bX_t-\bY_{st}\|_F^2 & \nonumber\\
&= \mathrm{Tr}\big(\bU_t\bB\bU_s^\top (\bX_s \bX_s^\top)\bU_s\bB\bU_t^\top (\bX_t\bX_t^\top)\big) + |\Omega| & \nonumber\\
& \qquad- 2\sum\limits_{\{(i,j):(i,j)\in\Omega\}} 	x_{si}^\top\bU_s\bB\bU_t^\top x_{tj},&\label{eqn:linearObjective}
\end{flalign}
where $x_{si}$ represents the $i$-{th} column in $\bX_s$, $x_{tj}$ represents the $j$-{th} column in $\bX_t$, $\Omega$ is the set of row-column indices corresponding to entry value $1$ in $\bY_{st}$, and $\mathrm{Tr}(\cdot)$ denotes the trace of a matrix. The complexity of computing the first and third term in (\ref{eqn:linearObjective}) is $O(d^3+n_sd^2+n_td^2)$ and $O(nd+n_sd^2+n_td^2)$, respectively. Similarly, the computation cost of the gradient of the objective function in (\ref{eqn:bilingualsquareloss}) is also linear in $n$. 
Hence, our framework can  efficiently leverage information from all the negative samples. 

In the next section, we discuss a generalization of our approach to multilingual settings.

\subsection{Generalization to Multilingual Setting}\label{subsec:multilingual}
In this section, we propose a unified framework for learning mappings when bilingual dictionaries are  available for multiple language pairs. 
We formalize the setting as an undirected, connected graph $G(V,E)$, where each node represents a language and an edge represents the availability of a bilingual dictionary between the corresponding pair of languages.  Given all bilingual dictionaries corresponding to the edge set $E$, we propose to align the embedding spaces of  all languages in the node set $V$ and learn a common latent space for them. 


To this end, we \textit{jointly} learn an orthogonal transformation $\bU_i\in\O^d$ for every language $L_i$ and the  Mahalanobis metric $\bB\succ\bzero$. 
The latter is \textit{common} across all languages in the multilingual setu p and helps incorporate information across languages in the latent space. 
It should be noted that the transformation $\bU_i$ is employed for all the bilingual mapping problems in this graph associated with $L_i$. 
The transformation from $L_i$ to $L_j$ is given by $\bW_{ji}=\bU_j\bB\bU_i^\top$. 
Further, we are also able to obtain transformations between any language pair in the graph, even if a bilingual dictionary between them is not available. 


Let $\bX_{i}^j\in\R^{d\times m}$ be\footnote{For notational convenience, the number of unique words in every language in all their dictionaries is kept same ($m$).} 
the embeddings of the dictionary words of $L_i$ in the dictionary corresponding to edge $e_{ij}\in E$. Let $\bY_{ij}\in\{0,1\}^{m\times m}$ be the binary label matrix corresponding to the dictionary between $L_i$ and $L_j$.  The proposed optimization problem for multilingual setting is 
\begin{flalign}\label{eqn:multilingual}
\minop_{\substack{\bU_i\in\O^d\ \forall i\\ \bB\succ\bzero}}\ & \sum_{e_{ij}\in E} \frac{1}{|\Omega_{ij}|}\|(\bX_{i}^j)^\top\bU_i\bB\bU_j^\top \bX_{j}^i-\bY_{ij}\|_F^2\nonumber\\
& +  \lambda \|\bB\|_F^2.
\end{flalign}

We term our approach as \textbf{Geo}metry-aware \textbf{M}ultilingual \textbf{M}apping (GeoMM). 
We next discuss the optimization algorithm for solving the bilingual mapping problem (\ref{eqn:bilingualsquareloss}) as well as its generalization to the multilingual setting (\ref{eqn:multilingual}).

\section{Optimization Algorithm}
\label{sec:optimization}
The geometric constraints $\bU_s\in\O^d ,\bU_t\in\O^d$ and $\bB\succ\bzero$ in the proposed problems (\ref{eqn:bilingualsquareloss}) and (\ref{eqn:multilingual}) have been studied as smooth Riemannian manifolds, which are well explored topological spaces \citep{edelman98a}. The orthogonal matrices  $\bU_i$ lie in, what is popularly known as, the $d$-dimensional Orthogonal manifold. The space of $d\times d$ symmetric positive definite matrices $(\bB\succ\bzero)$ is known as the Symmetric Positive Definite manifold. 
The Riemannian optimization framework embeds such constraints into the search space and conceptually views the problem as an unconstrained problem over the manifolds. In the process, it is able to exploit the geometry of the manifolds and the symmetries involved in them. 
\citet{absil08a} discuss several tools to systematically optimize such problems. 
We optimize the problems (\ref{eqn:bilingualsquareloss}) and (\ref{eqn:multilingual}) using the Riemannian conjugate gradient algorithm \citep{absil08a,sato13a}. 

Publicly available toolboxes such as {Manopt} \citep{boumal14a}, {Pymanopt}  \citep{townsend16a} or ROPTLIB \citep{huang16a} have scalable off-the-shelf generic implementations of several Riemannian optimization algorithms. We employ Pymanopt in our experiments, where we only need to supply the objective function. 


\section{Experimental Settings}
\label{sec:exp_setup}

In this section, we describe the evaluation tasks, the datasets used, and the experimental details of the proposed approach.  


\noindent\textbf{Evaluation tasks}. We evaluate our approach on several tasks:

\bulletpoint To evaluate the quality of the bilingual mappings generated, we evaluate our algorithms primarily for the bilingual lexicon induction (BLI) task\ie word translation task and compare Precision@1 with previously reported state-of-the-art results on benchmark datasets \citep{dinu15a,artetxe16a,conneau18a}. 

\bulletpoint We also evaluate on the cross-lingual word similarity task using the SemEval 2017 dataset. 

\bulletpoint To ensure that quality of embeddings on monolingual tasks does not degrade, we evaluate the quality of our embeddings on the monolingual word analogy task \citep{artetxe16a}.

\bulletpoint To illustrate the utility of representing embeddings of multiple language in a single latent space, we evaluate our multilingual embeddings on the one-hop translation task\ie a direct dictionary between the source and target languages is not available, but the source and target languages share a bilingual dictionary with a pivot language. 


\noindent\textbf{Datasets}. For bilingual and multilingual experiments, we report results on the following widely used, publicly available datasets: 

\bulletpoint \textbf{VecMap}: This dataset was originally made available by \citet{dinu15a} with subsequent extensions by other researchers \citep{artetxe17a,artetxe18a}.  It contains bilingual dictionaries from English (en) to four languages: Italian (it), German (de), Finnish (fi) and Spanish (es). 
The detailed experimental settings for this BLI task can be found in \citet{artetxe18b}. 

\bulletpoint \textbf{MUSE}: This dataset was originally made available by \citet{conneau18a}. 
It contains bilingual dictionaries from English to many languages such as Spanish (es), French (fr), German (de), Russian (ru), Chinese (zh), and \textit{vice versa}. 
The detailed experimental settings for this BLI task can be found in \citet{conneau18a}. This dataset also contains bilingual dictionaries between several other European languages, which we employ in multilingual experiments. 


\noindent\textbf{Experimental settings of GeoMM}. We select the regularization hyper-parameter $\lambda$ from the set $\{10,10^2,10^3,10^4\}$ by evaluation on a validation set created out of the training dataset. For inference, we use the {(normalized)} latent space representations of embeddings ($\bB^{\frac{1}{2}} \bU_i^\top x$) to compute similarity between the embeddings. For inference in the bilingual lexicon induction task, we employ the Cross-domain Similarity Local Scaling (CSLS) similarity score \citep{conneau18a} in nearest neighbor search, unless otherwise mentioned. CSLS has been shown to perform better than other methods in mitigating the \textit{hubness} problem \citep{dinu15a} for search in high dimensional spaces.  

While discussing experiments, we denote our bilingual mapping algorithm (Section~\ref{subsec:bilingual}) as {\algname} and its generalization to the multilingual setting (Section~\ref{subsec:multilingual}) as {\algname$_{\mathrm{multi}}$}.  Our code is available at \url{https://github.com/anoopkunchukuttan/geomm}. 


\section{Direct Translation: Results and Analysis}
\label{sec:dir_trans}

In this section, we evaluate the performance of our approach on two tasks: bilingual lexicon induction and cross-lingual word similarity. We also perform ablation tests to understand the effect of major sub-components of our algorithm. We verify the monolingual performance of the mapped embeddings generated by our algorithm. 

\begin{table*}[t]\centering
\setlength{\tabcolsep}{5pt}
{\footnotesize
\centering
\begin{tabular}{lrrrrrrrrrrr}
\toprule
\lhead{Method} & \thead{en-es}  & \thead{es-en} & \thead{en-fr}  & \thead{fr-en} & \thead{en-de}  & \thead{de-en} & \thead{en-ru}  & \thead{ru-en} & \thead{en-zh}  & \thead{zh-en} & \thead{avg.} \\
\cmidrule(lr){1-1}
\cmidrule(lr){2-11}
\cmidrule(lr){12-12}
{\text{\underline{Supervised}}} & & & & & & & & & & & \\
\textbf{\algname}    & $81.9$ & $85.5$ & $82.1$ & $\mathbf{84.2}$ & $74.9$ & $\mathbf{76.7}$ & $\mathbf{52.8}$ & $\mathbf{67.6}$ & ${49.1}$ & $\mathbf{45.3}$ & $\mathbf{70.0}$\\ 
\textbf{\algname}$_{\mathrm{multi}}$  & $81.0$ & $\mathbf{85.7}$ & $81.9$ & ${83.9}$ & $75.1$ & ${75.7}$ & ${51.7}$ & ${67.2}$ & $\mathbf{49.4}$ & ${44.9}$ & ${69.7}$\\
Procrustes         & $81.4$ & $82.9$ & $81.1$ & $82.4$ & $73.5$ & $72.4$ & ${51.7}$ & $63.7$ & $42.7$ & $36.7$ & $66.9$ \\ 
MSF-ISF            & $79.9$ & $82.1$ & $80.4$ & $81.4$ & $73.0$ & $72.0$ & $50.0$ & $65.3$ & $28.0$ & $40.7$ & $65.3$ \\ 
MSF                  & $80.5$ & $83.8$ & $80.5$ & $83.1$ & $73.5$ & $73.5$ & $50.5$ & $67.3$ & $32.3$ & $43.4$ & $66.9$\\ 
MSF$_\mu$       & $80.3$ & $84.0$ & $80.7$ & $83.9$ & $73.1$ & $74.7$ & $\times$ & $\times$ & $\times$ & $\times$ & $-$ \\ 
{\text{\underline{Unsupervised}}} & & & & & & & & & & & \\
SL-unsup           & $82.3$ & $84.7$  & $82.3$  & $83.6$  & $75.1$  & $74.3$  & $49.2$  & $65.6$  & $0.0$  & $0.0$ & $59.7$ \\  
Adv-Refine$^{*}$         & $81.7$ & $83.3$ & $82.3$ & $82.1$ & $74.0$ & $72.2$ & $44.0$ & $59.1$ & $32.5$ & $31.4$ & $64.3$ \\ 
\citet{grave18}$^{*}$  & $\mathbf{82.8}$ & $84.1$ & $\mathbf{82.6}$ & $82.9$ & $\mathbf{75.4}$ & $73.3$ & $43.7$ & $59.1$ & $-$ & $-$ & $-$ \\ 
\citet{hoshen18}$^{*}$  & ${82.1}$ & $84.1$ & ${82.3}$ & $82.9$ & ${74.7}$ & $73.0$ & $47.5$ & $61.8$ & f.c. & f.c. & $-$ \\ 
\citet{chen18a}$^{*}$ & $82.5$ & $83.7$ & $82.4$ & $81.8$ & $74.8$ & $72.9$ & $-$ & $-$ & $-$ & $-$ & $-$\\ 
\bottomrule
\end{tabular}
}
\caption{Precision$@1$ for BLI on the MUSE dataset. Some notations: (a) `$-$' implies the original paper does not report result for the corresponding language pair, (b) `f.c.' implies the original paper reports their algorithm failed to converge, (c) `$\times$' implies that we could not run the authors' code successfully for the language pairs, and (d) `{}$^*$' implies the results of the algorithm are reported in the original paper. 
The remaining results were obtained with the official implementation from the authors. }\label{table:bilingual-conneau}
\end{table*}
\begin{table}[t]\centering
\setlength{\tabcolsep}{5.9pt}
\centering
{\footnotesize
\begin{tabular}{lrrrrr}
\toprule
\lhead{Method} & \thead{en-it}  & \thead{en-de} & \thead{en-fi}  & \thead{en-es} &  \thead{avg.}\\
\cmidrule(lr){1-1}
\cmidrule(lr){2-5}
\cmidrule(lr){6-6}
{\text{\underline{Supervised}}} & & & & &  \\
\textbf{\algname}  & $48.3$ & $\mathbf{49.3}$ & $\mathbf{36.1}$ & $\mathbf{39.3}$ & $\mathbf{43.3}$\\
\textbf{\algname}$_{\mathrm{multi}}$  & $\mathbf{48.7}$ & ${49.1}$ & ${36.0}$ & ${39.0}$ & ${43.2}$\\
Procrustes & $44.9$ & $46.5$ & $33.5$ & $35.1$ & $40.0$ \\ 
MSF-ISF  & $45.3$ & $44.1$ & $32.9$ & $36.6$ & $39.7$\\ 
MSF  & $47.7$ & $47.5$ & $35.4$ & $38.7$ & $42.3$\\ 
MSF$_\mu$ & ${48.4}$ & $47.7$ & $34.7$ & $38.9$ & $42.4$ \\ 
GPA & $45.3$ & $48.5$ & $31.4$ & $-$ & $-$ \\ 
CCA-NN & $38.4$ & $37.1$ & $27.6$ & $26.8$ & $32.5$\\ 
{\text{\underline{Unsupervised}}} & & & & &  \\ 
SL-unsup & $48.1$ & $48.2$ & $32.6$ & $37.3$ & $41.6$\\ 
Adv-Refine & $45.2$ & $46.8$ & $0.4$ & $35.4$ & $31.9$\\ 
\bottomrule
\end{tabular}
}
\caption{Precision$@1$ for BLI on the VecMap dataset. 
The results of MSF-ISF, SL-unsup, CCA-NN \citep{faruqui14a}, and Adv-Refine are reported by  \citet{artetxe18b}. 
CCA-NN employs nearest neighbor retrieval procedure. 
The results of GPA are reported by \citet{kementchedjhieva18a}. 
}\label{table:bilingual-artetxe}
\end{table}
\begin{table}\centering
\centering
{\footnotesize
\begin{tabular}{lrrrr}
\toprule
\lhead{Method} & \thead{en-it}  & \thead{en-de} & \thead{en-fi}  & \thead{en-es}\\
\cmidrule(lr){1-1}
\cmidrule(lr){2-5}
\textbf{{\algname}}  & $\mathbf{48.3}$ & $\mathbf{49.3}$ & $\mathbf{36.1}$ & $\mathbf{39.3}$ \\
\midrule
(1) $\bW\in\R^{d\times d}$ & $45.4$ & $47.9$ & $35.4$ & $37.5$ \\
(2) $\bW=\bB$      & $26.3$ & $26.3$ & $19.5$ & $21.2$ \\
(3) $\bW=\bU_t\bU_s^\top$ & $13.2$  & $16.0$  & $8.8$  & $11.8$ \\
(4) Targt space inf. & $45.5$ & $47.8$ & $35.0$ & $37.9$ \\
(5) Regression  & $46.8$ & $43.3$ & $33.9$ & $35.4$ \\
\bottomrule
\end{tabular}
}
\caption{Ablation test results: Precision$@1$ for BLI on the VecMap dataset.}\label{table:ablation}
\end{table}
\subsection{Bilingual Lexicon Induction (BLI)}


We compare {\algname} with the best performing supervised methods. We also compare with unsupervised methods as they have been shown to be competitive with supervised methods. The following baselines are compared in the BLI experiments. 

\bulletpoint Procrustes: the bilingual mapping is learned by solving the orthogonal Procrustes problem \citep{xing15a,artetxe16a,smith17a,conneau18a}. 



\bulletpoint MSF: the Multi-Step Framework proposed by \citet{artetxe18a}, with CSLS retrieval. It improves upon the original system (MSF-ISF) by \citet{artetxe18a}, which employs inverted softmax function (ISF) score for retrieval. 


\bulletpoint Adv-Refine: unsupervised adversarial training approach, with bilingual dictionary refinement \citep{conneau18a}. 

\bulletpoint SL-unsup: state-of-the-art self-learning (SL) unsupervised method \citep{artetxe18b}, employing structural similarity of the embeddings. 

We also include results of the correction algorithm proposed by \citet{doval18a} on the MSF results (referred to as MSF$_\mu$).
In addition, we also include results of several recent works \citep{kementchedjhieva18a,grave18,chen18a,hoshen18} on MUSE and VecMap datasets, which are reported in the original papers. 



\noindent\textbf{Results on MUSE dataset}: Table \ref{table:bilingual-conneau} reports the results 
on the MUSE dataset. 
We observe that our algorithm {\algname} outperforms all the supervised baselines. 
{\algname} also obtains significant improvements over unsupervised approaches.

The performance of the multilingual extension, {\algname}$_{\mathrm{multi}}$, is almost equivalent to the bilingual {\algname}. This means that in spite of multiple embeddings being jointly learned and represented in a common space, its performance is still better than existing bilingual approaches. Thus, our multilingual framework is quite robust since languages from diverse language families have been embedded in the same space. This can allow downstream applications to support multiple languages without performance degradation.  Even if bilingual embeddings are represented in a single vector space using a pivot language, the embedding quality is inferior compared to {\algname}$_{\mathrm{multi}}$. We discuss more multilingual experiments in Section~\ref{sec:indir_trans}.

\noindent\textbf{Results on VecMap dataset}: Table \ref{table:bilingual-artetxe} reports the results on the  VecMap dataset. We observe that {\algname} obtains the best performance in each language pair, surpassing state-of-the-art results reported on this dataset. GeoMM also outperforms GPA \citep{kementchedjhieva18a}, which also learns bilingual embeddings in a latent space.

\subsection{Ablation Tests}

We next study the impact of different components of our framework by varying one component at a time. The results of these tests on VecMap dataset are shown in Table \ref{table:ablation} and are discussed below. 

\noindent\textbf{(1)}  \textbf{Classification with unconstrained $\bW$}. We learn the transformation $\bW$ directly as follows: 
\begin{equation}
\minop_{\bW\in\R^{d\times d}}  \lambda \|\bW\|_F^2 + \|\bX_s^\top\bW^\top\bX_t - \bY_{st}  \|_F^2. 
\end{equation}
The performance drops in this setting compared to {\algname}, underlining the importance of the proposed factorization and the latent space representation.  In addition, the proposed factorization helps {\algname} generalize to the multilingual setting ({\algname}$_{\mathrm{multi}}$). 
Further, we also observe that the overall performance of this simple classification based model is better than recent supervised approaches such as Procrustes, MSF-ISF \citep{artetxe18a}, and GPA \citep{kementchedjhieva18a}. This suggests that a classification model is better suited for the BLI task. 

Next, we look at both components of the factorization. 

\noindent\textbf{(2)} \textbf{Without language specific rotations}. We enforce $\bU_s=\bU_t=\bI$ in (\ref{eqn:bilingualsquareloss}) for \algname\ie $\bW=\bB$. We observe a significant drop in performance, which highlights the need for aligning the feature space of different languages. 

\noindent\textbf{(3)} \textbf{Without similarity metric}. We enforce $\bB=\bI$ in (\ref{eqn:bilingualsquareloss}) for \algname\ie $\bW=\bU_t\bU_s^\top$. It can be observed that the results are poor, which underlines the importance of a suitable similarity metric in the proposed classification model. 

\noindent\textbf{(4)} \textbf{Target space inference}. We learn $\bW=\bU_t\bB\bU_s^\top$ by solving (\ref{eqn:bilingualsquareloss}), as in {\algname}. During the retrieval stage, the similarity between embeddings is computed in the target space\ie given embeddings $x$ and $z$ from the source and target languages, respectively, we compute the similarity of the (normalized) vectors $\bW x$ and $z$. 
It should be noted that {\algname} computes similarity of $x$ and $z$ in the latent space\ie it computes the similarity of the (normalized) vectors $\bB^{\frac{1}{2}} \bU_s^\top x$ and $\bB^{\frac{1}{2}} \bU_t^\top z$, respectively. 
We observe that  inference in the target space degrades the performance. This shows that the latent space representation captures useful information and allows {\algname} to obtain much better accuracy.  

\noindent\textbf{(5)}  \textbf{Regression with proposed factorization}. We pose BLI as a regression problem, as done in  previous approaches, by employing the following loss function: $\|\bU_t\bB\bU_s^\top\bX_s - \bX_t  \|_F^2$. 
We observe that its performance is worse than the classification baseline ($\bW\in\R^{d\times d}$). 
The classification setting directly models the similarity score via the loss function, and hence corresponds with inference more closely. 
This result further reinforces the observation made in the first ablation test.
%

To summarize, the proposed modeling choices are better than the alternatives compared in the ablation tests.


\subsection{Cross-lingual Word Similarity}

The results on the cross-lingual word similarity task using the SemEval 2017 dataset \citep{collados17} are shown in Table~\ref{table:crosslingual-wordsim-semeval17}. We observe that {\algname} performs better than Procrustes, MSF, and the SemEval 2017 baseline {NASARI} \citep{collados16}. It is also competitive with Luminoso\_run2 \citep{speer17}, the best reported system on this dataset. It should be noted that {NASARI} and {luminoso\_run2} use additional knowledge sources like BabelNet and ConceptNet.


\begin{table}[t]\centering
{\small
\centering
\begin{tabular}{lrrr}
\toprule
\lhead{Method} & \thead{en-es}  & \thead{en-de} & \thead{en-it}\\
\cmidrule(lr){1-1}
\cmidrule(lr){2-4}
NASARI   & $0.64$ & $0.60$ & $0.65$\\
Luminoso\_run2  & $0.75$ & $0.76$ & $0.77$\\
\midrule
Procrustes & $0.72$ & $0.72$ & $0.71$\\
MSF  & $0.73$ & $0.74$ & $0.73$\\
\citet{joulin18b}  & $0.71$ & $0.71$ & $0.71$\\
\textbf{{\algname}}  & $0.73$ & $0.74$ & $0.74$\\
\bottomrule
\end{tabular}
}
\caption{Pearson correlation coefficient for the SemEval 2017 cross-lingual word similarity task.} \label{table:crosslingual-wordsim-semeval17}
\end{table}
\begin{table}[t]\centering
{\small
\centering
\begin{tabular}{lr}
\toprule
\lhead{Method} & \thead{Accuracy (\%)} \\
\midrule 
Original English embeddings & $76.66$ \\
Procrustes  & $76.66$ \\ 
MSF & $76.59$ \\
\textbf{{\algname}} &  $75.21$\\
\bottomrule
\end{tabular}
}
\caption{Results on the monolingual word analogy task.}\label{table:monolingual-wordana}
\end{table}


\subsection{Monolingual Word Analogy}

Table \ref{table:monolingual-wordana} shows the results on the English monolingual analogy task after obtaining it$\rightarrow$en mapping on the VecMap dataset \cite{mikolov13b,artetxe16a}. 
We observe that there is no significant drop in the monolingual performance by the use of non-orthogonal mappings compared to monolingual embeddings as well as other bilingual embeddings (Procrustes and MSF). 

%
%

\section{Indirect Translation: Results and Analysis}
\label{sec:indir_trans}

In the previous sections, we have established the efficacy of our approach for bilingual mapping problem when a bilingual dictionary between the source and target languages is available. 
We also showed that our proposed multilingual generalization (Section~\ref{subsec:multilingual}) performs well in this scenario. 
In this section, we explore if our multilingual generalization is beneficial when a bilingual dictionary is not available between the source and the target, in other words, \textit{indirect translation}. For this evaluation, our algorithm learns a \textit{single model} for various language pairs such that word embeddings of different languages are transformed to a common latent space. 

\subsubsection*{Evaluation Task: One-hop Translation}
We consider the BLI task from language $L_{\mathrm{src}}$ to language $L_{\mathrm{tgt}}$ in the absence of a bilingual lexicon between them. We, however, assume the availability of lexicons for $L_{\mathrm{src}}$-$L_{\mathrm{pvt}}$ and $L_{\mathrm{pvt}}$-$L_{\mathrm{tgt}}$, where $L_{\mathrm{pvt}}$ is a {\it pivot} language. 


As baselines, we adapt any supervised bilingual approach (Procrustes, MSF, and the proposed {\algname}) to the one-hop translation setting by considering their following variants: 

\bulletpoint \textbf{Composition} ($\mathrm{cmp}$): Using the given bilingual approach, we learn the $L_{\mathrm{src}}\rightarrow L_{\mathrm{pvt}}$ and $L_{\mathrm{pvt}}\rightarrow L_{\mathrm{tgt}}$ transformations as $\bW_1$ and $\bW_2$, respectively. Given an embedding $x$ from  $L_{\mathrm{src}}$, the corresponding embedding in $L_{\mathrm{tgt}}$ is obtained by a composition of the transformations\ie $\bW_2\bW_1 x$. This is equivalent to computing the similarity of $L_{\mathrm{src}}$ and $L_{\mathrm{tgt}}$ embeddings in the $L_{\mathrm{pvt}}$ embedding space.  Recently, \citet{smith17b} explored this technique with the Procrustes algorithm. 

\bulletpoint \textbf{Pipeline} ($\mathrm{pip}$): Using the given bilingual approach, we learn the $L_{\mathrm{src}}\rightarrow L_{\mathrm{pvt}}$ and $L_{\mathrm{pvt}}\rightarrow L_{\mathrm{tgt}}$ transformations as $\bW_1$ and $\bW_2$, respectively. Given a word embedding $x$ from  $L_{\mathrm{src}}$, we infer its translation embedding $z$ in $L_{\mathrm{pvt}}$. 
Then, the corresponding embedding of $x$ in $L_{\mathrm{tgt}}$ is $\bW_2 z$. 


As discussed in Section~\ref{subsec:multilingual}, our framework allows the flexibility to jointly learn the common latent space of multiple languages, given bilingual dictionaries of multiple language pairs. Our multilingual approach, {\algname}$_{\mathrm{multi}}$, views this setting as a graph with three nodes $\{L_{\mathrm{src}},L_{\mathrm{tgt}},L_{\mathrm{pvt}}\}$ and two edges $\{L_{\mathrm{src}}$-$L_{\mathrm{pvt}},L_{\mathrm{pvt}}$-$L_{\mathrm{tgt}}\}$ (dictionaries). 

\begin{table}[t]
\setlength{\tabcolsep}{4.9pt}
\centering
{\small
\begin{tabular}{lrrrr}
\toprule
\lhead{Method } & \thead{fr-it-pt} & \thead{it-de-es} & \thead{es-pt-fr} & \thead{avg.} \\
\midrule
SL-unsup                                                 & $74.1$ & $86.4$ & $84.6$ & $81.7$ \\ 
\midrule
{\textbf{\underline{Composition}}} \\
Procrustes                                               & $74.2$ & $81.9$ & $82.5$ & $79.5$ \\
MSF                                                     & $75.3$ & $81.9$ & $82.7$ & $80.0$ \\
{\algname}                            & $77.7$ & $84.1$ & $84.3$ & $82.0$ \\
\midrule
{\textbf{\underline{Pipeline}}} \\
Procrustes                                              & $72.5$ & $61.6$ & $79.9$ & $71.3$ \\ 
MSF                                                      & $75.9$ & $64.5$ & $82.5$ & $74.3$ \\
{\algname}                            & $75.9$ & $62.5$ & $81.7$ & $73.4$ \\ 
\midrule
\textbf{\algname}$_\mathrm{multi}$ & $\mathbf{80.1}$ & $\mathbf{86.8}$ & $\mathbf{85.6}$ & $\mathbf{84.2}$ \\
\bottomrule
\end{tabular}
}
\caption{Indirect translation: Precision$@1$ for BLI.}\label{table:onehop_results}
\end{table}

\subsubsection*{Experimental Settings}

We experiment with the following one-hop translation cases: (a) fr-it-pt, (b) it-de-es, and (c) es-pt-fr (read the triplets as $L_{\mathrm{src}}$-$L_{\mathrm{pvt}}$-$L_{\mathrm{tgt}}$). The training/test dictionaries and the word embeddings are from the MUSE dataset. In order to minimize direct transfer of information from $L_{\mathrm{src}}$ to $L_{\mathrm{tgt}}$, we generate  $L_{\mathrm{src}}$-$L_{\mathrm{pvt}}$ and $L_{\mathrm{pvt}}$-$L_{\mathrm{tgt}}$ training dictionaries such that they do not have any $L_{\mathrm{pvt}}$ word in common. The training dictionaries have the same size as the $L_{\mathrm{src}}$-$L_{\mathrm{pvt}}$ and $L_{\mathrm{pvt}}$-$L_{\mathrm{tgt}}$ dictionaries provided in the MUSE dataset while the test dictionaries have $1\,500$ entries. 


\begin{table*}[t]\centering
\setlength{\tabcolsep}{4pt}
{\small
\centering
\begin{tabular}{lrrrrrrrrrrrrr}
\toprule
\lhead{Method} & \thead{en-es}  & \thead{es-en} & \thead{en-fr}  & \thead{fr-en} & \thead{en-de}  & \thead{de-en} & \thead{en-ru}  & \thead{ru-en} & \thead{en-zh}  & \thead{zh-en} & \thead{en-it}  & \thead{it-en} & \thead{avg.} \\
\cmidrule(lr){1-1}
\cmidrule(lr){2-11}
\cmidrule(lr){12-13}
\cmidrule(lr){14-14}
RCSLS              & $\mathbf{84.1}$ & $86.3$ & $\mathbf{83.3}$ & $84.1$ & $\mathbf{79.1}$ & $76.3$ & $\mathbf{57.9}$ & $67.2$ & $45.9$ & $\mathbf{46.4}$ & $45.1$ & $38.3$ & $66.2$\\
\textbf{\algname}    & $81.9$ & $85.5$ & $82.1$ & $84.2$ & $74.9$ & $\mathbf{76.7}$ & $52.8$ & $67.6$ & $\mathbf{49.1}$ & $45.3$ & $48.3$ & $41.2$ & $65.8$\\ 
\textbf{{\algname}$_\mathrm{semi}$}  & $82.7$ & $\mathbf{86.7}$ & $82.8$ & $\mathbf{84.9}$ & $76.4$ & $\mathbf{76.7}$ & $53.2$ & $\mathbf{68.2}$ & $48.5$ & ${46.1}$ & $\mathbf{50.0}$ & $\mathbf{42.6}$ & $\mathbf{66.6}$\\ 
\bottomrule
\end{tabular}
}
\caption{Comparison of {\algname} and {\algname}$_\mathrm{semi}$ with RCSLS \citep{joulin18b}. Precision$@1$ for BLI is reported. The results of RCSLS are reported in the original paper. The results of language pairs en-it and it-en are on the VecMap dataset, while others are on the MUSE dataset.}\label{table:bilingual-joulin}
\end{table*}
\subsubsection*{Results and Analysis}

Table \ref{table:onehop_results} shows the results of the one-hop translation experiments. 
We observe that {\algname}$_{\mathrm{multi}}$ outperforms pivoting methods ($\mathrm{cmp}$ and $\mathrm{pip}$) built on top of MSF and Procrustes for all language pairs. It should be noted that pivoting may lead to cascading of errors in the solution, whereas learning a common embedding space jointly mitigates this disadvantage. This is reaffirmed by our observation that {\algname}$_{\mathrm{multi}}$ performs significantly better than  {\algname} (cmp) and {\algname} (pip). 
 


Since unsupervised methods have been shown to be competitive with supervised methods, they can be an alternative to pivoting. Indeed, we observe that the unsupervised method {SL-unsup} is better than the pivoting methods though it used no bilingual dictionaries. On the other hand, {\algname}$_{\mathrm{multi}}$ is better than the unsupervised methods too. It should be noted that the unsupervised methods use much larger vocabulary than {\algname}$_{\mathrm{multi}}$ during the training stage. 

We also experimented with scenarios where some words from $L_{\mathrm{pvt}}$ occur in both $L_{\mathrm{src}}$-$L_{\mathrm{pvt}}$ and $L_{\mathrm{pvt}}$-$L_{\mathrm{tgt}}$ training dictionaries. In these cases too, we observed that {\algname}$_{\mathrm{multi}}$ perform better than other methods. We have not included these results due to space constraints.

\section{Semi-supervised {\algname}}\label{sec:semi_sup}

In this section, we discuss an extension of {\algname}, which benefits from  unlabeled data. For the bilingual mapping problem, unlabeled data is available in the form of vocabulary lists for both the source and target languages. Existing unsupervised and semi-supervised techniques  \citep{artetxe17a,artetxe18b,joulin18b,hoshen18} have an iterative refinement procedure that employs the vocabulary lists to augment the dictionary with positive or negative mappings.  

Given a seed bilingual dictionary, we implement a bootstrapping procedure that iterates over the following two steps until convergence:
\begin{enumerate}
\item Learn the {\algname} model by solving the proposed formulation (\ref{eqn:bilingualsquareloss}) with the current bilingual dictionary.
\item Compute a new bilingual dictionary  from the vocabulary lists, using the (current) {\algname} model for retrieval.  The seed dictionary along with this new dictionary is used in the next iteration. 
\end{enumerate}
In order to keep the computational cost low, we restrict the vocabulary list to $k$ most frequent words for both the languages \citep{artetxe18b,hoshen18}. In addition, we perform bidirectional dictionary induction \citep{artetxe18b,hoshen18}. 
We track the model's performance on a validation set to avoid overfitting and use it as a criterion for convergence of the bootstrap procedure. 

We evaluate the proposed semi-supervised {\algname} algorithm (referred to as \textbf{{\algname}$_{\mathrm{semi}}$}) on the bilingual lexicon induction task on MUSE and VecMap datasets. The bilingual dictionary for training is split $80/20$ into the seed dictionary and the validation set. We set   $k=25\,000$, which works well in practice. 

We compare {\algname}$_{\mathrm{semi}}$ with RCSLS, a recently proposed state-of-the-art semi-supervised algorithm by \citet{joulin18b}. RCSLS directly optimizes the CSLS similarity score \citep{conneau18a}, which is used during retrieval stage for {\algname}, among other algorithms. On the other hand, {\algname}$_{\mathrm{semi}}$ optimizes a simpler classification based square loss function (refer Section \ref{subsec:bilingual}). In addition to the training dictionary, RCSLS uses the full vocabulary list of the source and target languages ($200\,000$ words each) during training.

The results are reported in Table \ref{table:bilingual-joulin}. We observe that the overall performance of {\algname}$_{\mathrm{semi}}$ is slightly better than RCSLS. In addition, our supervised approach {\algname} performs slightly worse than RCSLS, though it does not have the advantage of learning from unlabeled data, as is the case for RCSLS and  {\algname}$_{\mathrm{semi}}$. 
We also notice that {\algname}$_{\mathrm{semi}}$ improves upon {\algname} in almost all language pairs. 

\begin{table}[t]\centering
\setlength{\tabcolsep}{5.9pt}
\centering
{\footnotesize
\begin{tabular}{lrrrrr}
\toprule
\lhead{Method} & \thead{en-it}  & \thead{en-de} & \thead{en-fi}  & \thead{en-es} &  \thead{avg.}\\
\cmidrule(lr){1-1}
\cmidrule(lr){2-5}
\cmidrule(lr){6-6}
\textbf{\algname}  & $48.3$ & ${49.3}$ & ${36.1}$ & ${39.3}$ & ${43.3}$\\
\textbf{\algname}$_{\mathrm{semi}}$  & $\mathbf{50.0}$ & $\mathbf{51.3}$ & $\mathbf{36.2}$ & $\mathbf{39.7}$ & $\mathbf{44.3}$\\
\bottomrule
\end{tabular}
}
\caption{Precision$@1$ for BLI on the VecMap dataset. 
}\label{table:bilingual-vecmap-semi}
\end{table}
We also evaluate {\algname}$_{\mathrm{semi}}$ on the VecMap dataset. The results are reported in Table~\ref{table:bilingual-vecmap-semi}. 
To the best of our knowledge, {\algname}$_{\mathrm{semi}}$ obtains state-of-the-art results on the VecMap dataset.

\section{Conclusion and Future Work}
\label{sec:conclusion}

In this work, we develop a framework for learning multilingual word embeddings by aligning the embeddings for various languages in a common space and inducing a Mahalanobis similarity metric in the common space. We view the translation of embeddings from one language to another as a series of geometrical transformations and jointly learn the language-specific orthogonal rotations and the symmetric positive definite matrix representing the Mahalanobis metric. Learning such transformations can also be viewed as learning a suitable common latent space for multiple languages.
We formulate the problem in the Riemannian optimization framework, which models the above transformations efficiently. 

We evaluate our bilingual and multilingual algorithms on the bilingual lexicon induction and the cross-lingual word similarity tasks. The results show that our algorithm outperforms existing approaches on multiple datasets. In addition, we demonstrate the efficacy of our multilingual algorithm in a one-hop translation setting for bilingual lexicon induction, in which a direct dictionary between the source and target languages is not available. The semi-supervised extension of our algorithm shows that our framework can leverage unlabeled data to obtain further improvements. 
Our analysis shows that the combination of the proposed transformations, inference in the induced latent space, and modeling the problem in classification setting allows the proposed approach to achieve state-of-the-art performance. 
In future, an unsupervised extension to our approach can  be explored. Optimizing the CSLS loss function \citep{joulin18b} within our framework can be investigated to address the hubness problem. We plan to work on downstream applications like text classification, machine translation, \textit{etc.}, which may potentially benefit from the proposed latent space representation of multiple languages by sharing annotated resources across languages. 


\bibliography{main-1563-Jawanpuria}
\bibliographystyle{acl_natbib}

\end{document}